\documentclass[runningheads]{llncs}
\pdfoutput=1
\usepackage{graphicx}
\usepackage{amsmath,amssymb} 
\usepackage{color}
\usepackage{epsfig}
\usepackage{xspace}
\usepackage[width=122mm,left=12mm,paperwidth=146mm,height=193mm,top=12mm,paperheight=217mm]{geometry}
\usepackage{caption}
\usepackage{subcaption}
\makeatletter
\DeclareRobustCommand\onedot{\futurelet\@let@token\@onedot}
\def\@onedot{\ifx\@let@token.\else.\null\fi\xspace}

\DeclareMathOperator*{\argmin}{\arg\!\min}

\def\ie{\emph{i.e}\onedot}

\def\etal{\emph{et al}\onedot}
\makeatother

\begin{document}
\pagestyle{headings}
\mainmatter
\def\ECCV16SubNumber{***}  

\title{Person Re-identification for Real-world Surveillance Systems} 



\author{Furqan M. Khan \and Fran\c{c}ois Br\'{e}mond}
\institute{INRIA Sophia Antipolis - M\'{e}diterran\'{e}e \\ 
2004 Route des Lucioles, Sophia Antipolis \\
\small{\{furqan.khan $\lvert$ francois.bremond\}@inria.fr} }

\maketitle

\begin{abstract}
Appearance based person re-identification in a real-world video surveillance system with non-overlapping camera views is a challenging problem for many reasons. 
Current state-of-the-art methods often address the problem by relying on supervised learning of similarity metrics or ranking functions 
to implicitly model appearance transformation between cameras for each camera pair, or group, in the system.
This requires considerable human effort to annotate data (see Section~\ref{Sec:Intro:CostAnnotation}). 
Furthermore, 
the learned models are camera specific and not transferable from one set of cameras to another.
Therefore, the annotation process is required 
after every network expansion or camera replacement, which strongly limits their applicability.
Alternatively, we propose a novel modeling approach to harness complementary appearance information without supervised learning that 
significantly outperforms current state-of-the-art unsupervised methods on multiple benchmark datasets. 

\end{abstract}

\section{Introduction}
\label{Sec:Intro}
\begin{figure}[tbh]
\begin{center}
  \begin{subfigure}{0.19\linewidth}
    \centering
    \includegraphics[width=0.45\linewidth]{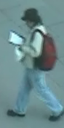}
    \includegraphics[width=0.45\linewidth]{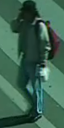}
  \caption{~}
   \end{subfigure}
  \begin{subfigure}{0.19\linewidth}
    \centering
    \includegraphics[width=0.33\linewidth]{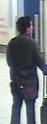}
    \includegraphics[width=0.33\linewidth]{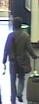}
  \caption{~}
   \end{subfigure}
  \begin{subfigure}{0.19\linewidth}
    \centering
    \includegraphics[width=0.45\linewidth]{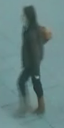}
    \includegraphics[width=0.45\linewidth]{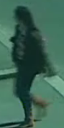}
   \caption{~}
   \end{subfigure}
  \begin{subfigure}{0.19\linewidth}
    \centering
    \includegraphics[width=0.35\linewidth]{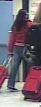}
    \includegraphics[width=0.45\linewidth]{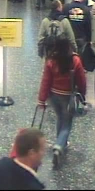}
  \caption{~}
   \end{subfigure}
\end{center}
\vspace{-6mm}
   \caption{Appearance variation of an individual in one track}
\label{fig:appearance}
\vspace{-7mm}
\end{figure}

The goal of person re-identification (Re-ID) is to identify a person at distinct times, locations, or in different camera views. The problem often arises in the context of search for individuals or 
long term tracking in a multi-camera visual surveillance system.
In a real-world system, Re-ID of a person is very challenging due to significant variation in an individual's 
appearance due to changes in camera properties, lighting, viewpoint and pose. In contrast, inter-person appearance similarity is generally very high in absence of biometric cues, such as face or iris, 
due to low resolution imaging or viewpoint (Fig~\ref{fig:appearance}).
Occlusions may impede visibility, and because a Re-ID system is driven by automatically acquired person tracks in practice, the individual may be only partially 
visible or not centered. These are significant challenges for appearance based Re-ID algorithms which often formulate the task as a matching problem for individuals' corresponding appearance {\em signatures}.

The Re-ID process is often divided into two stages: i) representing each person using his appearance signature acquired from 
image(s), and ii) sorting candidate matches using a similarity metric or a ranking function of appearance signatures. 
Re-ID task is classified as either \emph{single-shot} or \emph{multi-shot} based on the number of images available to learn each signature. 
For Re-ID in video surveillance systems, it is possible to use multiple images of a person to learn his appearance signature by grouping images using an off-the-shelf tracking algorithm. 
Therefore, this paper focuses on the multi-shot case.

Having multiple images case can be useful in learning robust appearance signatures; 
however, trivial solutions, such as averaging information from multiple images, get affected by variance in a person's appearance. 
Therefore, optimally combining information from multiple images into one signature and 
defining suitable metric for that signature representation is a non-trivial problem.

Recent trend in the literature is to overcome weakness of low-level features in handling complex Re-ID scenarios by using supervised machine 
learning techniques to adapt a similarity metric or a ranking function for a set of cameras
~\cite{Ahmed15,BakECCV2012,Chen15,Dikmen2010,Kostinger2012,Liao15,Liu15,Shen15,Su15,Wang2014,Zhao2014}. 
Although significant improvement is possible, 
high annotation effort (Sec~\ref{Sec:Intro:CostAnnotation}) associated with supervised learning makes it unsuitable for real-world systems. 
Alternatively, this paper focuses on improving signature \emph{representation} for multi-shot scenario and avoiding supervised learning for scalability.

The approach in this paper uses a rich representation for signatures, called Multi Channel Appearance Mixture.
The representation is novel to multi-shot Re-ID and capable of more accurately encoding a person's multi-modal appearance 
using Gaussian Mixture Models and multiple features. 
The idea is to judiciously consider variance in a person's appearance and independence of features to 
find suitable number and description of mixture components to compactly represent his signature. 
Finally, similarity between two signatures is defined as a combination of f-divergence and 
Collaborative Coding (\cite{ZhangSparse2011}) based distance that does not require supervised learning and 
hence is real-world systems friendly.
%

The components of our approach, such as GMMs and the low-level features, are not novel; instead, novelty is in the means they are 
convened together to address the task at hand through careful consideration of multi-shot Re-ID problem and a person's appearance.  
It is due to this improved way of assembling different components and representing multi-shot signatures that our method outperforms state-of-the-art unsupervised approaches, 
and most supervised approaches, on multiple datasets - SAIVT-SoftBio~\cite{Alina}, PRID2011~\cite{Hirzer2011}, and iLIDS-VID~\cite{Wang2014}.

\subsection{Annotation Effort for Supervised Model Learning}
\label{Sec:Intro:CostAnnotation}
Approaches such as~\cite{Ahmed15,BakECCV2012,Chen15,Dikmen2010,Kostinger2012,Liao15,Liu15,Shen15,Su15,Wang2014,Zhao2014} 
either learn a metric, like Mahalanobis distance, or a ranking function using supervised machine learning. 
Two types of annotation are needed for person tracks: bounding boxes, and unique identities. 
Even though automated person detection and tracking can be used to aid with marking bounding boxes, existing methods are
far from perfect. Therefore, fragmentation and ID-switches are quite common and human effort is required to  
resolve these issues and assign unique identities. This work is quite tedious and data is noisy; consequently, most Re-ID methods train models using human annotated tracks.

Most of the above methods learn one metric or ranking function per camera pair, except for~\cite{Su15}, which uses Multiple Task Learning framework 
to train multiple multi-class classifiers for a group of cameras together. In either case, considerable human effort is required to annotate data. 
In the pairwise case, given $N$ cameras, $N(N-1)/2$ pairwise models are required. Considering, that in a 
typical real-world scenario not all persons pass through all the cameras due to non-overlapping camera views and multiple entries and exits, 
one may have to annotate $2p$ samples (tracks in multi-shot case) to train one model with $p$ persons. Therefore, a total of $O(N^2p)$ samples have to be annotated. 
That is, to train each model with $100$ persons for a network of $10$ cameras, approximately $9,900$ track samples are required, 
which is quite expensive. Furthermore, as the models are camera specific, adding or replacing one camera requires a minimum of 
another $10\times100$ samples; therefore, the annotation cost is recurrent.

\section{Related Work and Contribution}
\label{Sec:RelatedWork}
Considerable effort has been dedicated in the past to improve both aspects, signature modeling and metric/rank-function design, of Re-ID process  
through inventive feature design (\cite{SlawekBTP14,Bazzani2010,SDALF2010,Gray2008,Hirzer2012,Karanam15,LiuPR2012,Liu2012,Prosser2010,Schwartz09,Tuzel06,Wang2007,Zeng15,Zhao13,Zheng2011}) 
and/or employing supervised learning (\cite{Ahmed15,BakECCV2012,Chen15,Dikmen2010,Kostinger2012,Liao15,Liu15,Shen15,Su15,Wang2014,Zhao2014,LiLBDM15}). 
Majority of these methods address single-shot scenario and use either a concatenated vector, or an ordered set, of multiple features to represent 
a person's signature. 

\begin{figure}[h]
\begin{center}
  \includegraphics[width=0.45\linewidth]{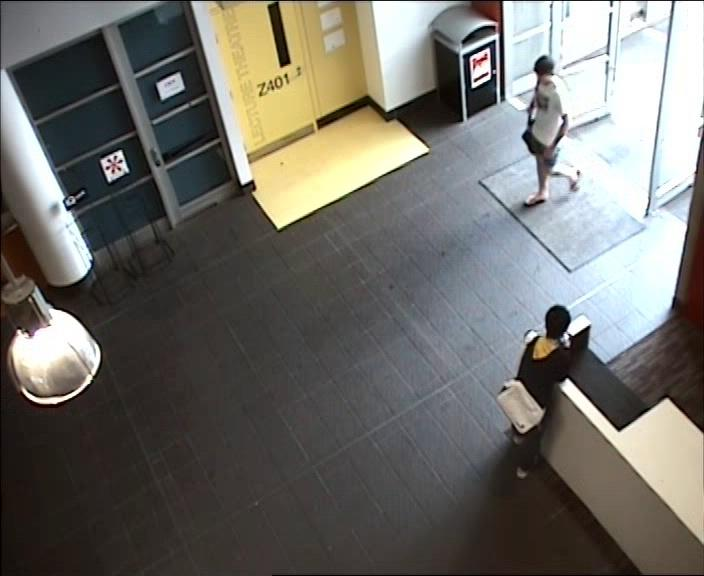}
  \includegraphics[width=0.45\linewidth]{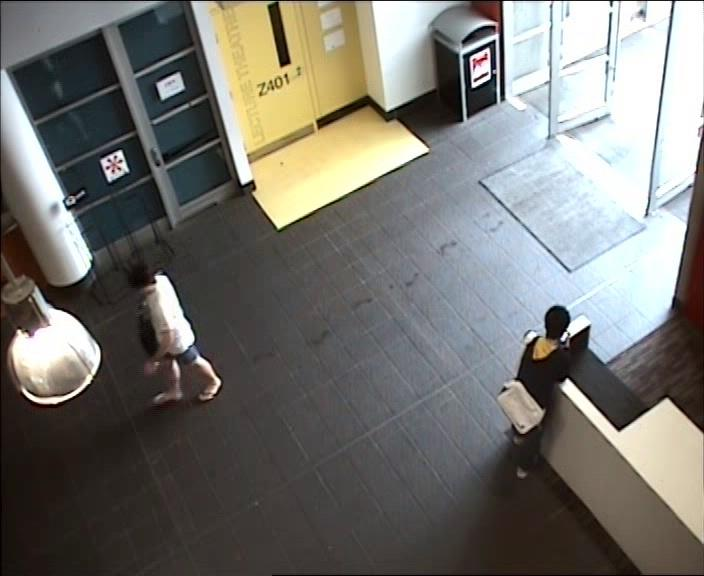}
\end{center}
   \caption{Appearance variation of an individual in one track}
\label{fig:demo}
\end{figure}

Single-shot methods are often trivially extended to perform multi-shot task by either representing a multi-shot signature as a set of image descriptors or by their 
average~\cite{Hirzer2011,Zeng15,LiLBDM15,LiRSCNN13}. 
The latter strategy makes incorrect assumption about uni-modality of appearance when the existing features are not sufficiently robust to deal 
with intra-person variance. Therefore, their performance is generally low.  
On the other hand, for the former representation, set similarity metrics, such as 
RSCNN~\cite{LiRSCNN13}, LBDM~\cite{LiLBDM15}, and CRC-S~\cite{Zeng15}, have been used to improve multi-shot signature matching. 
Evaluation of these metrics for signatures computed for tracks,~\ie sets with large cardinality, is computationally expensive. 
This necessitates limiting the number of image descriptors in the set. 
Additionally, LBDM assumes there is only one track per person in the query and gallery sets, which is often not true due to track fragmentation.

Uninformed random sampling used by~\cite{Zeng15,LiLBDM15,LiRSCNN13} to limit set cardinality is prone to losing valuable information, specially if images belonging 
to a certain appearance mode are very few. 
For instance in Fig~\ref{fig:demo}, as the person in white shirt walks across the room, the number of image samples in the track 
from brighter region of the room will be significantly fewer than from the darker region. 
A fixed size random sample may miss all the brighter samples, while sampling at regular interval makes the set cardinality grow linearly with time; 
both of which are undesirable. 
Conversely, the proposed approach uses feature information to compress signature size while retaining significantly more information. 

In our knowledge, only few approaches,~\cite{Slawek14,Bazzani2010,SDALF2010,Wang2014,Liu15}, adequately capture multi-modality of appearance for Re-ID. 
Understandably, they outperform unimodal methods despite their theoretical shortcomings. 
Bazzani~\etal~\cite{Bazzani2010} and Farenzena~\etal~\cite{SDALF2010} use appearance cues to segment tracks but assume that appearance modes in different feature domains are aligned with HSV histogram. 
Intuitively, this assumption fails when features are independent or they intend to capture complementary information, which limits efficiency of multiple feature fusion. 
For example, shape features may not vary considerably with illumination change but color features would. 
On the other hand, Bak~\etal~\cite{Slawek14} use orientation cues, while Wang~\etal~\cite{Wang2014} and Liu~\etal~\cite{Liu15} use motion cues, to discover track segments for different appearance modes.
Both of them, however, ignore the effect of lighting and other factors on a person's appearance. 

We address these issues by: 
i) independently learning probability distribution of each feature as a multi-modal Gaussian represented as a Gaussian Mixture Model; and
ii) using variance of features as a cue to discover appearance modes instead of ``external'' ones like orientation or pose, because
most low-level features are \emph{not} robust to arbitrary transformations,
such as pose changes; thus variance based cues subsume pose and orientation cues. 
Further, by using GMMs we retain more information about appearance of a person 
that allows for better discrimination between persons with similar appearance. 
Moreover, unlike Liu~\etal, who learn one GMM per \emph{action unit} with fixed number of components on an additional training set, 
one GMM per track per feature with variable number of components and do not require any data for training.

In summary, we contribute towards solution of multi-shot Re-ID problem through MCAM representation of multi-shot signatures by:
\begin{itemize}
	\vspace{-2mm}
  \item Advocating to discover appearance modalities in the domain of each feature being used to preserve their complementary nature.
  \item Efficiently retaining and utilizing additional appearance information about a person through GMMs and suitable metrics to help resolve difficult cases.
  \item Using feature variance as a cue to discover and describe multiple modes of person appearance which makes learned signatures more robust to pose 
  illumination and viewpoint changes.
  \item Improving representation to avoid use of human involvement during model construction, which allows handling arbitrary number of persons and camera views.
	\vspace{-4mm}
\end{itemize}

\section{Multi Channel Appearance Mixtures}
\label{Sec:MCAM}

The objective of this paper is to address multi-shot Re-ID problem for multi-camera surveillance scenario, where the goal is to associate different tracks of a person in the same or different cameras. 
The Re-ID process is preceded by localization of different persons in space and time using a person detector, 
followed by linking of different detections into short term tracks using an object tracker.
As person detection and tracking are beyond the scope of Re-ID methods and this paper, we assume that some state-of-the-art detection and tracking method is used to create a query set $\bar{Q}$ and a gallery set $\bar{G}$ of person \emph{tracks}. We, however, make no assumption 
about the source of the two sets. That is, the sets may correspond to two cameras, a set of cameras, or one camera. 
However, for ease of discussion, we may often refer to inter-camera association scenario, as it is more common in multi-camera video surveillance. 
Further, there is no limit on the number of tracks that belong to a particular person in one set because it is probable that the track of a person is fragmented. 
\subsection{Signature Representation}
\label{Sec:representation}
Under multi-camera surveillance, a person may exhibit multiple modes of appearance in one track due to variation in illumination, viewpoint, and/or pose. 
Therefore, it is important that multi-modality of appearance be handled explicitly. 
However, the number of modes of a person and corresponding image frames are not known apriori. Therefore, both problems of ``mode discovery'' - finding number of modes and corresponding frames, 
and ``mode description'' - appearance description using low-level features, need to be solved. 
Our strategy is to use variance in low-level feature descriptors as a cue to solve both problems simultaneously. 
This strategy can be easily realized by representing a person's appearance as a multi-modal Gaussian distribution of features and learning its parameters so that each mode has a low variance 
and is far from other modes. These objectives can be easily achieved by using Gaussian Mixture Models and Expectation Maximization algorithm for learning parameters. 

We define Multi Channel Appearance Mixture (MCAM) as a representation for multi-shot signatures that combines multiple appearance 
models (Gaussian mixtures) corresponding to different low-level feature \emph{channels}. 
The representation is extensible to any number and type of low-level features because for each feature, its corresponding appearance model is 
learned independent of others.

Given a track $t = \{I^t_n : n=1:N_t\} \in \bar{Q} \cup \bar{G}$ of length $N_t$ and a set of features $F$, 
the corresponding MCAM signature $\tilde{t} = \{ \mathcal{M}^t_f : f \in F\}$ is defined as 
a set of appearance models $\mathcal{M}^t_f$, one for each feature $f$. 
In turn, each feature appearance model defines density of feature $f$ for track $t$ using a multivariate Gaussian Mixture Model (GMM) representation, 
$\mathcal{M}^t_f = \{ \pi^t_{f,k}, \mathcal{G}^t_{f,k} : k = 1:K^t_f\}$ with $K^t_f$ components, where $\pi^t_{f,k}$ is the prior probability of the $k^{th}$ Gaussian 
component $\mathcal{G}^t_{f,k} \sim \mathcal{N}(\boldsymbol{\mu}^t_{f,k},\boldsymbol{\Sigma}^t_{f,k})$ having mean $\boldsymbol{\mu}^t_{f,k}$ 
and covariance $\boldsymbol{\Sigma}^t_{f,k}$. 

\subsubsection{Appearance learning}
Parameters of each appearance mixture $\mathcal{M}^t_f$ are estimated independently for each track $t$ and feature $f$. \
Given the set of feature descriptors $S^t_f=\{\boldsymbol{s}^t_{f,n} : n=1:N_t\}$ corresponding to images $\{I^t_n : n=1:N_t\}$ and feature $f$, by ignoring temporal relationship among images, 
the parameters of each appearance model (GMM) can be easily estimated using EM algorithm. 
However, in practice, we trade-off accuracy with computational cost by using k-means algorithm to first obtain component means and estimate 
covariance matrix only after k-means algorithm has converged. This is equivalent to the assumption that all Gaussian components share a fixed covariance matrix during their mean estimation. 

Since tracks have variable length and features are independent, each track and feature may require different number of components to correctly represent the appearance. 
Thus, the number of components $K^t_f$ cannot be fixed apriori. 
A model selection technique, such as Bayesian Information Criterion or Alkaline Information Criterion, can be used to automatically 
discover suitable number of components for each signature $\tilde{t}$ and feature $f$. However, we found that the following simple regularized formulation 
that trades average cluster distortion with the number of components of appearance mixture yields 
satisfactory results. 
\begin{equation}
 K^t_f = \argmin_{K=1:K_{max}} J(S^t_f,K) + h(K) \label{eq:oDistortion}
\end{equation}
\begin{equation}
 J(S^t_f,K) = \min_{\substack{\mu_{i=1:K},\\c(\boldsymbol{s}^t_{f,1}),...,c(\boldsymbol{s}^t_{f,{N_t}})}} 
			\frac{1}{K}\sum_n^{N_t} \left\lVert\boldsymbol{s}^t_{f,n}-\boldsymbol{\mu}_{c(\boldsymbol{s}^t_{f,n})}\right\rVert_2 \label{eq:distortion}
\end{equation}%
where, $K_{max}$ is the maximum number of components allowed, and the function $J(.)$ represents minimum average cluster distortion, $h(K)$ is the penalty function, 
$\mu_i$ is the mean descriptor for the $i^{th}$ cluster and $c(.)$ is the cluster assignment function that maps an appearance descriptor to its cluster number. 
The formulation favors fewer components, if $h(K)$ is an increasing function of $K$. During experiments we found that $h(K) = sqrt(K)$ gives satisfactory performance.

For computational efficiency, we use kmeans++~\cite{Arthur07} that 
allows k-means to converge faster. Given error bounds in~\cite{Arthur07}, we run k-means 
for a maximum of 10 iterations and achieve good results. Therefore, running k-means multiple times does not create a bottleneck even for moderately long tracks. 
Furthermore, covariance matrices, which are restricted to be diagonal for efficiency, are only computed after optimal number of components have been found. 
%

\subsubsection{Feature descriptors}
A number of low-level features have been proposed for Re-ID task in the past. 
Color based features~\cite{SDALF2010} work reasonably well in low density datasets, as opposed to crowded ones, because the probability of people wearing same color clothing is low. 
On the other hand, shape based features are robust to illumination changes but struggle to exhibit enough discriminative power by themselves when resolution of images is low. 
Therefore, we use complementary shape and color information to represent a person`s signature. 
Our approach is capable of incorporating a number of features; however, for experimentation we used the following three features that capture 
complementary appearance information: 
\begin{itemize}
\vspace{-1mm}
 \item {\bf Color spatio-histogram} (CSH) as described in~\cite{Zeng15}; however, we use $30$ bin histograms separately for each of the color channels in Lab color space. 
 \item {\bf Histogram of oriented gradients} (HOG)~\cite{Dalal05} over $8$ bins of signed orientation with $L_1$ normalization.
 \item {\bf Brownian covariance of features} (BCov)~\cite{SlawekBTP14} using intensities and their gradients (both magnitudes and orientations) for each of the RGB channel and the pixel locations $x$ and $y$. 
\vspace{-1mm}
\end{itemize}

Before computing any of the features, we crop out image of the person, re-scale it to a fixed size window of $w\times h$ pixels and apply histogram equalization to the $L$ channel of the $Lab$ color image 
to minimize illumination variance. Each image is then subdivided into a number of rectangular overlapping regions, denoted by set $\boldsymbol{R}$. Features are extracted from each of the sub-windows. 
Corresponding features are concatenated into one vector $\boldsymbol{s}^t_{f,n}$ to represent appearance of the $n^{th}$ image of track $t$ in channel $f$. 
We project the covariance features onto the tangent plane~\cite{SlawekBTP14} before concatenation. 
The features can be computed independently in parallel and hence using multiple features isn't more computationally expensive than using one feature. 
%

\section{Similarity Metric for MCAM}
\label{Sec:SimilarityMetric}

We define similarity between two signatures $\tilde{q}$ and $\tilde{g}$ as a sum of two complementary similarity measures:
i) $L_2$-Riemannian similarity, $Sim_{LR}(\tilde{q},\tilde{g})$, and 
ii) Collaborative Representation Coding based similarity $Sim_{CRCS}(\tilde{q},\tilde{g})$.
\begin{equation}
	Sim(\tilde{q},\tilde{g}) = Sim_{LR}(\tilde{q},\tilde{g}) + Sim_{CRCS}(\tilde{q},\tilde{g})
\end{equation}

\subsection{$L_2$-Riemannian similarity}
We represent each signature using a set of GMMs; 
therefore, Jeffrey's divergence (symmetric KL-divergence) or Hellinger distance can be used to compute distance between two Gaussian components and define the overall signature similarity based on it.
However, Abou-Moustafa~\etal~\cite{AbouMoustafa12} noted that for Gaussian densities, both Jeffrey's divergence and Hellinger distance 
can be factorized into terms corresponding to distance between the first and the second order moments,~\ie mean and covariance, and  
the term corresponding to distance between covariances can be replaced with Riemannian metric for symmetric positive definite matrices to yield a modified 
$\alpha$-weighted distance measure while maintaining metric properties as follows:
%
\begin{equation}
	\label{eq:distGauss}
	d(\mathcal{G}_1,\mathcal{G}_2;\alpha) = (1-\alpha)(\boldsymbol{u}^T\boldsymbol{\Psi}\boldsymbol{u})^{\frac{1}{2}} + 
			\alpha d_{\mathcal{R}}(\boldsymbol{\Sigma_1},\boldsymbol{\Sigma_2})
\end{equation}
where, $\mathcal{G}_1\sim\mathcal{N}(\boldsymbol{\mu}_1,\boldsymbol{\Sigma}_1)$ and 
$\mathcal{G}_2\sim\mathcal{N}(\boldsymbol{\mu}_2,\boldsymbol{\Sigma}_2)$ are two multivariate Gaussian distributions 
with mean and covariance $\boldsymbol{\mu}_1, \boldsymbol{\Sigma}_1$ and $\boldsymbol{\mu}_2, \boldsymbol{\Sigma}_2$, respectively; 
$\boldsymbol{u}=\boldsymbol{\mu}_1 - \boldsymbol{\mu}_2$ is the difference of mean vectors; 
$\boldsymbol{\Psi} = \boldsymbol{\Sigma}_1^{-1} + \boldsymbol{\Sigma}_2^{-1}$, in case of Jeffrey's divergence, or 
$\boldsymbol{\Psi} = (\frac{1}{2}\boldsymbol{\Sigma}_1 + \frac{1}{2}\boldsymbol{\Sigma}_2)^{-1}$, for Hellinger distance; 
$\alpha \in (0,1)$ controls weight of the two terms; 
and $d_R(,)$ is the Riemannian metric between the two covariance matrices defined as follows:

\begin{equation}
	\label{eq:distRiem}
 d_R(\boldsymbol{\Sigma}_1,\boldsymbol{\Sigma}_2) = \left(\sum_{p=1}^P log^2{\lambda_p}\right)^{\frac{1}{2}}
\end{equation}
where, $dig(\lambda_1, \lambda_2,...,\lambda_P) = \boldsymbol{\Lambda}$ is the generalized eigenvalue matrix for the generalized eigenvalue problem: 
$\boldsymbol{\Sigma}_1\boldsymbol{V} = \boldsymbol{\Lambda\Sigma}_2\boldsymbol{V}$, 
and $\boldsymbol{V}$ is the column matrix of its generalized eigenvectors. 
Eq.~\ref{eq:distRiem} can be efficiently solved for diagonal covariances.

Note that the first term in Eq.~\ref{eq:distGauss} measures Mahalanobis distance between Gaussian means and it is possible to 
completely decouple the two terms of Eq.~\ref{eq:distGauss} by choosing any arbitrary positive semi-definite 
matrix $\boldsymbol{\Psi}$. Optimal matrix can be estimated using supervised metric learning techniques; 
however, due to high annotation cost we avoid supervised learning and instead replace the term with $L_2$ distance between the Gaussian means,
~\ie we set $\boldsymbol{Psi}=\boldsymbol{I}$.
This gives us the following $\alpha$-weighted definition of distance between two Gaussians: 
\begin{equation}
	\label{eq:distLR}
 d_{LR}(\mathcal{G}_1,\mathcal{G}_2; \alpha) = (1-\alpha)\left\|\boldsymbol{\mu_1} - \boldsymbol{\mu_2}\right\|_2 + 
		\alpha d_{\mathcal{R}}(\boldsymbol{\Sigma_1},\boldsymbol{\Sigma_2})
\end{equation}

Using Eq.~\ref{eq:distLR}, we define channel-wise distance $D_{LR}(\mathcal{M}^q_f,\mathcal{M}^g_f)$ between two appearance mixtures 
$\mathcal{M}^q_f$ and $\mathcal{M}^g_f$ as the minimum distance between a Gaussian component $\mathcal{G}_i \in \mathcal{M}^q_f$ 
and a component $\mathcal{G}_j \in \mathcal{M}^g_f$.
\begin{equation}
	\label{eq:distLRChannel}
	D_{LR}(\mathcal{M}^q_f,\mathcal{M}^g_f) = \min_{\mathcal{G}_i \in \mathcal{M}^q_f, \mathcal{G}_j \in \mathcal{M}^g_f} d_{LR}(\mathcal{G}_i, \mathcal{G}_j; \alpha_{ij})
\end{equation}

The relative weight parameter $\alpha_{ij}$ in Equation~\ref{eq:distLRChannel} is determined using corresponding prior probabilities of Gaussian components in each appearance 
mixture. However, we limit the influence of the covariance component based on number of frames used to construct a signature,knowing that it is more important that two appearance mixtures agree on their means and that too few frames may result in poor covariance 
estimation.  We estimate the upper limit $\alpha_{max}$ on the influence of covariance component 
and the value of $\alpha_{ij}$ for a particular pair of Gaussian components $\mathcal{G}_i$ and $\mathcal{G}_j$ as:
\begin{eqnarray}
 \alpha_{max} = \min(a,\min(N_q,N_g)/b) \label{eq:alphamax} \\
 \alpha_{ij} = \min(\alpha_{max}, (\bar{\pi}_i + \bar{\pi}_j)/2)
\end{eqnarray}
where, $a$ defines the global upper limit on the influence of covariance component; 
$b$ controls the rate at which $\alpha_{max}$ can increase as a function of minimum of number of images used to create signatures $\tilde{q}$ and $\tilde{g}$; 
$N_q$, $N_g$ are the number of images used to create signatures $\tilde{q}$ and $\tilde{g}$ respectively; and $\bar{\pi}_i$, $\bar{\pi}_j$ are the max-normalized 
prior probabilities of Gaussians $\mathcal{G}_i$ and $\mathcal{G}_j$ respectively.

The channel-wise distances for a query signature $\tilde{q}$ are then converted to similarity by applying a Gaussian kernel after 
normalizing with the maximum distance between the query and a gallery signature. The overall similarity between a query signature $\tilde{q}$ and 
a gallery signature $\tilde{g}$ is then defined as:
\begin{equation}
	\label{eq:simLR}
	Sim_{LR}(\tilde{q}, \tilde{g}) = \sum_{f \in F} exp\left(-\gamma_f^{-1}\left(\bar{D}_{LR}(\mathcal{M}^q_f,\mathcal{M}^g_f) - \beta_f\right)^2\right)
\end{equation}
where, $\bar{D}_{LR}(\mathcal{M}^q_f,\mathcal{M}^g_f) = D_{LR}(\mathcal{M}^q_f,\mathcal{M}^g_f)/\max_{\bar{g} \in G} D_{LR}(\mathcal{M}^q_f,\mathcal{M}^{\bar{g}}_f)$ is 
max normalized over gallery set $G$, 
$\beta_f = \min_{\bar{q} \in Q} \bar{D}_{LR}(\mathcal{M}^{\bar{q}}_f,\mathcal{M}^g_f)$ is defined as the minimum normalized distance for the gallery signature $\tilde{g}$ 
from a signature in query set $Q$ and $\gamma_f = 0.33*range_{\bar{q} \in Q} \bar{D}_{LR}(\mathcal{M}^{\bar{q}}_f,\mathcal{M}^g_f)$ is one-third of the range 
of the distance over query set $Q$, implying that similarity goes to 0 at the max distance.

\subsection{Collaborative Representation Coding based similarity}
Recently, Collaborative Representation Coding (CRC) has been 
used to compute similarity between two multi-shot signatures~\cite{Zeng15}. 
The idea is to encode a query signature 
using the dictionary $\boldsymbol{\mathcal{D}}$ constructed from \emph{all} gallery signatures $\tilde{g} \in \tilde{G}$, 
such that the reconstruction error is minimized. Then ability of a gallery signature to represent the query signature is measured relative to 
optimal coding.

Even though this has shown significant improvement over Euclidean set based measures, 
it is not easy to adapt this distance to include component variances of GMM without paying significant computational cost. 
Therefore, we use CRC to only measure discrepancy between the mean vectors of different Gaussian components. 
Specifically, we adapt CRC-S from~\cite{Zeng15} to compute distance between two appearance mixtures $\mathcal{M}^q_f$ and $\mathcal{M}^g_f$ 
corresponding to feature $f$ as follows (for clarity, we drop subscript $f$ from notation):

First, given an appearance mixture $\mathcal{M}^g$ we construct a corresponding matrix 
$\mathcal{D}^g = [\mu^g_1~...~\mu^g_{K^g}]$ using component means. Then the dictionary matrix 
$\boldsymbol{\mathcal{D}} = [\mathcal{D}^1~\mathcal{D}^2~...~\mathcal{D}^{|G|}]$ is constructed using matrices for gallery 
signatures $\{\mathcal{D}^g : g \in G\}$. 
Afterwards, mean $\mu^q_i$ of $i^{th}$ Gaussian component $\mathcal{G}^q_i$ of query signature $\tilde{q}$ is encoded using dictionary matrix $\boldsymbol{\mathcal{D}}$ and weight vector $\boldsymbol{\rho}$ by optimizing the following objective: 
\begin{equation}
	\label{eq:crcs-prob}
 \arg\min_{\boldsymbol{\rho}} \left\|\mu^q_i - \boldsymbol{\mathcal{D}}\boldsymbol{\rho}\right\|_2 + \delta\left\|\boldsymbol{\rho}\right\|_2
\end{equation}

Problem in Equation~\ref{eq:crcs-prob} has a closed form solution:
\begin{equation}
 \label{eq:optRho}
 \boldsymbol{\rho} = (\boldsymbol{\mathcal{D}}^T\boldsymbol{\mathcal{D}} + \delta\boldsymbol{I})^{-1}\boldsymbol{\mathcal{D}}^T \mu^q_i
\end{equation}

Next, encoding vector $\boldsymbol{\rho}_g$ corresponding to signature $\tilde{g}$ is extracted from $\boldsymbol{\rho}$ and 
is used to define the distance between $i^{th}$ mixture component of $\tilde{q}$ and mixture model $\mathcal{M}^g$ of signature $\tilde{g}$ as a combination of residual error
when encoding $\mu^q_i$ using 
only the dictionary $\mathcal{D}^g$ corresponding to $\tilde{g}$ with weights $\boldsymbol{\rho}_g$ and the regularization term 
for coding vector $\boldsymbol{\rho}_g$.
\begin{equation}
\label{eq:distCrcs}
 d_{CRCS}(\mathcal{G}^q_i, \mathcal{M}^g) = \left\|\mu^q_i - \mathcal{D}^g\boldsymbol{\rho}_g\right\|_2 - \eta\left\|\boldsymbol{\rho}_g\right\|_2 
\end{equation}

Finally, distance between two appearance mixtures $\mathcal{M}^g_f$ and $\mathcal{M}^g_f$ for corresponding feature $f$ is defined as weighted sum of distance between 
the $i^th$ component of appearance mixture $\mathcal{M}^q_f$ and the appearance mixture $\mathcal{M}^g_f$ with corresponding prior probabilities $\pi^q_{f,i}$ as weights.
\begin{equation}
	D_{CRCS}(\mathcal{M}^g_f, \mathcal{M}^g_f) = \sum_{i=1:K^q_f} \pi^q_{f,i} d_{CRCS}(\mathcal{G}^q_{f,i}, \mathcal{M}^g_f) 
\end{equation}

CRCS distance between two signatures for each feature channel $f$ is converted into similarity using a similar process as described above for 
$L_1$-Riemannian distance (Equation~\ref{eq:simLR}). The only difference is that the two components of CRCS are max normalized over gallery separately before combination. 

\begin{equation}
	\label{eq:simLR}
	Sim_{CRCS}(\tilde{q}, \tilde{g}) = \sum_{f \in F} exp\left(-\bar{\gamma}_f^{-1}\left(\bar{D}_{CRCS}(\mathcal{G}^q_f,\mathcal{G}^g_f) - \bar{\beta}_f\right)^2\right)
\end{equation}
where $\bar{D}_{CRCS}(\mathcal{G}^q_f,\mathcal{G}^g_f)$, $\bar{\beta}_f$ and $\bar{\gamma}_f$ are defined similarly as above for $L_1$ Riemannian similarity.

\section{Evaluation}
\subsection{Implementation details}
There are four parameters related to similarity computation: $a$, $b$ in Eq.~\ref{eq:alphamax}, $\delta$ in Eq.~\ref{eq:optRho} and $\eta$ in Eq.~\ref{eq:distCrcs}. 
These and all other parameters related to features and signature
are fixed once for all datasets. $a=0.33$ and $b=100$ control maximum and slope of $\alpha_{max}$. We found that performance is not very sensitive to $a \in (0.33,0.5)$ and $b \in (50,100)$. 
Following~\cite{Zeng15}, $\delta = 1$ and $\eta$ is set to $0.55/0.45$ and 
the two components in Eq.~\ref{eq:distCrcs} are combined after normalization. 
Finally, the maximum number of mixture components $K_{max}$ is set to $K_{max} = \max(5, 0.1N_t)$, 
where $N_t$ is the length of track $t$. 
This allows for maximum number of components to vary with the length of track. Remember that this is the maximum number of components, 
exact number of components are discovered automatically.
For feature descriptors we re-scale 
all images to $64 \times 192$ pixels. Each image is then sub-divided into $|\boldsymbol{R}|=33$ overlapping regions of $32 \times 32$ pixels with 
$16$ pixels overlap. 

\noindent\textbf{Computation Time:} On a single core CPU, for iLIDS-VID dataset with average track length of 73, computing appearance model for HOG, BCov and CSH features take $1.6$, $3$, and $7$ seconds, respectively, on average per signature.
Computing $LR$ distance between two signatures take $\sim 7msec$ and computing $CRCS$ distance takes $\sim 230msec$ on average. 
Note that Re-ID, unlike detection and tracking, is not necessarily real-time. 
It is often run on-demand after tracks are acquired. Therefore, above times are quite reasonable for a practical system.

\vspace{-3mm}
\subsection{Datasets and experimental setup}
\vspace{-2mm}
Although there are many datasets available for evaluation of Re-ID methods, however, only a few are suitable for multi-shot Re-ID scenario. 
For experiments, we selected SAIVT-SoftBio~\cite{Alina}, PRID 2011~\cite{Hirzer2011}, and iLIDS-VID~\cite{Wang2014}.
However, since our approach is agnostic to the fact that {\em gallery} is constructed from one camera or multiple and that it does not require any training, all three datasets 
can be viewed as one large dataset. For each dataset, performance is reported using \emph{rank-N} recognition rates averaged over 10 trials.

\noindent\textbf{SAIVT-SoftBio dataset: }
SAIVT-SoftBio is collected from 8 cameras with non-overlapping views and provides the most realistic scenario for multi-shot Re-ID task 
due to multiple entry and exit points. 
The dataset consists of tracks of 152 persons. For evaluation, we used experimental setup of~\cite{Slawek14},~\ie 
we evaluate our approach pair-wise on all 56 possible camera pairs and report average results. 

\noindent\textbf{PRID 2011 dataset: }
PRID 2011 dataset consists of tracks from two cameras. The dataset is challenging due to data imbalance and high color inconsistency between both cameras. 
Tracks of $200$ and $749$ people are available for Camera A and Camera B, respectively. 
Tracks have variable lengths between $5$ - $675$ images. We experimented under two settings. First, to evaluate different aspects - features and metrics - of our model, 
we use entire dataset and experimental setup of~\cite{Hirzer2011},~\ie we use all $200$ persons from Camera A as query set and all $749$ person from Camera B as gallery set. 
Second, for fair comparison with competing methods, we used experimental setup of~\cite{Wang2014},~\ie we only considered people visible from both cameras and having at least $21$ images. 
The data is then equally and randomly divided into train and test sets, even though our method does not require any training. 

\noindent\textbf{iLIDS-VID dataset: }
iLIDS-VID is extracted from iLIDS MCTS dataset. It consists of 600 tracks of 300 people collected from non-overlapping cameras at an airport. The dataset is very challenging 
due to high amount of occlusion and low resolution. Similar to PRID 2011 dataset, we report results under two experimental setups.
First, we use all $300$ person to evaluate different aspects of our approach. Data from Camera B is used for query and from Camera A for gallery. 
Second, for fair comparison with others, similar to~\cite{Wang2014}, we equally and randomly divide data into train and test sets and evaluate our method.

%
\vspace{-3mm}
\subsection{Results and discussion}
\vspace{-2mm}
\subsubsection{Comparison of different features.}
To compare different features, we applied our method with only one feature at a time and compare the performance with complete multi feature model.  
Table~\ref{tab:featcomp} shows rank-$n$ recognition rates for MCAM Model with only 
color (\emph{CSH}), shape (\emph{HOG}) and texture ({\em BCov}) features, and MCAM approach combining all three features.

 \begin{table}[t]
 \centering
 \begin{tabular}{| l | c | c | c | c || c | c | c | c || c | c | c | c |}
 \hline
  & \multicolumn{4}{c||}{SAIVT-SoftBio} & \multicolumn{4}{c||}{PRID2011} & \multicolumn{4}{c|}{iLIDS-VID} 
   \\ \cline{2-13}
 Feature & ~r=1~ & ~r=5~ & ~r=10~ & ~r=20~ & ~r=1~ & ~r=5~ & ~r=10~ & ~r=20~ & ~r=1~ & ~r=5~ & ~r=10~ & r=20 
   \\ \hline 
    CSH & 25.0 & 49.4 & 61.7 & 75.4 & 15.5 & 33.0 & 40.5 & 49.5 & 17.3 & 43.0 & 51.7 & 62.3 
      \\ \hline 
    HOG & 26.5 & 44.0 & 54.9 & 67.6 & 29.0 & 52.0 & 61.0 & 70.0 & 26.3 & 48.0 & 58.7 & 69.3 
      \\ \hline 
    BCov & 21.2 & 43.0 & 59.0 & 74.0 & 17.0 & 33.0 & 42.5 & 52.0 & 21.3 & 40.7 & 51.7 & 63.7 
      \\ \hline 
    MCAM & \textbf{32.8} & \textbf{55.5} & \textbf{67.3} & \textbf{79.1} & \textbf{37.0} & \textbf{56.9} & \textbf{68.0} & \textbf{76.5} & \textbf{34.0} & \textbf{58.3} & \textbf{67.0} & \textbf{77.0} 
   \\ \hline 
 \end{tabular}
 \caption{Comparison of low-level features using recognition rate (\%) at different ranks \emph{r} on SAIVT-SoftBio, PRID2011 and iLIDS-VID datasets.}
 	\label{tab:featcomp}
 	\vspace{-10mm}
 \end{table}

It is evident from Table~\ref{tab:featcomp} that \emph{HOG} works better 
on all datasets. We believe that this is because \emph{PRID2011} has significant color disparity between cameras, and \emph{iLIDS-VID} has additional shape information provided by the luggage carried by persons. 
However, on \emph{SAIVT-SoftBio}, \emph{CSH} performs better than \emph{HOG}, except for rank-1. Finally, combining all the features result in 
significantly improved performance on all the datasets. This shows that the representation is capable of taking advantage of complementary information captured by different features.

\vspace{-4mm}
\subsubsection{Comparison of different metrics.}
Similarity between two signatures is based on a combination of $CRCS$ and $L_2$-Riemannian ($LR$) similarities. 
In order to assess significance of each similarity measure on Re-ID performance, we first applied our approach using only 
$LR$ similarity and then using only $CRCS$ based similarity. Finally, the two similarity measures are combined as explained in Sec.~\ref{Sec:SimilarityMetric}. 
All three features were used for each experiment. 
 \begin{table}[b]
 \vspace{-5mm}
 \centering
 \begin{tabular}{| l | c | c | c | c || c | c | c | c || c | c | c | c |
 	}
 \hline
  & \multicolumn{4}{c||}{SAIVT-SoftBio} & \multicolumn{4}{c||}{PRID2011} & \multicolumn{4}{c|}{iLIDS-VID} 
   \\ \cline{2-13}
 Metric~ & ~r=1~ & ~r=5~ & ~r=10~ & ~r=20~ & ~r=1~ & ~r=5~ & ~r=10~ & ~r=20~ & ~r=1~ & ~r=5~ & ~r=10~ & r=20 
   \\ \hline 
 LR & 26.6 & 50.6 & 62.8 & 77.3 & 24.0 & 44.5 & 53.5 & 63.5 & 23.2 & 45.3 & 55.7 & 67.3 
   \\ \hline 
 CRCS & 30.3 & 52.7 & 63.7 & 75.4 & 32.5 & 48.0 & 58.5 & 68.5 & 30.7 & 52.0 & 59.7 & 69.7 
   \\ \hline 
 MCAM & \textbf{32.8} & \textbf{55.5} & \textbf{67.3} & \textbf{79.1} & \textbf{37.0} & \textbf{56.9} & \textbf{68.0} & \textbf{76.5} & \textbf{33.0} & \textbf{57.3} & \textbf{65.0} & \textbf{75.7} 
   \\ \hline 
 \end{tabular}
 \caption{Comparison of similarity metrics using recognition rate (\%) at different ranks \emph{r} on SAIVT-SoftBio, PRID2011 and iLIDS-VID datasets. }
 	\label{tab:metric-comp}
\vspace{-8mm}
 \end{table}

Table~\ref{tab:metric-comp} shows rank-$n$ recognition rates on the datasets. 
Results indicate that CRCS is generally a better metric, however, it is significantly more ($\sim$10 times) more expensive to compute than LR. 
Combining both similarity metrics improve performancy by approximately 10\% - 15\% on each dataset. 
Performance gain on PRID2011 dataset is higher than others, which may be a consequence of high color disparity between two cameras that make complementary metric more meaningful. 
%

\vspace{-4mm}
\subsubsection{Comparison with state-of-the-art.} ~ \\
\noindent\textbf{SAIVT-SoftBio:}
We compared our approach against Re-ID by Viewpoint Cues (\emph{VCues})~\cite{Slawek14}, which uses viewpoint cues to learn multi-modal person signatures, 
and the baseline approach of~\cite{Slawek14} that randomly selects $10$ frames from each track to construct a signature. 
Since, \emph{VCues}, used only color information, we compared their method 
with our method when using only color channel appearance mixture (\emph{ColorAM}), as well as, when using all three features (\emph{MCAM}).
Our approach comprehensively outperforms their approach under both single and multi feature settings (Table~\ref{tab:sota-sb-unsup}). 
The boost in performance is a result of using feature cues 
to determine appearance modalities instead of orientation of persons. 

\begin{table}[t]
\vspace{-3mm}
    \centering
    \begin{tabular}{|l|*{4}{c|}}
    \hline
    Method~ & ~r=1~ & ~r=5~ & ~r=10~ & ~r=20~ \\ \hline 
    Baseline~\cite{Slawek14}~ & 7.1 & 21.5 & 35.1 & 52.0 \\ 
    VCues~\cite{Slawek14} & 22.8 & 41.5 & 53.8 & 67.7 \\ \hline
    ColorAM & 25.0 & 49.4 & 61.7 & 75.4 \\ 
     MCAM & \textbf{32.8} & \textbf{55.5} & \textbf{67.3} & \textbf{79.1} \\ \hline 
    \end{tabular}
\caption{Comparison of MCAM with state-of-the-art on SAIVT-SoftBio dataset using recognition rate (\%) at different ranks $r$}
	\label{tab:sota-sb-unsup}
	\vspace{-4mm}
\end{table}

\noindent\textbf{PRID 2011:}
As a reminder, we use experimental setup of~\cite{Wang2014} and only use partial dataset in evaluation for fair comparison with other methods. 
First, we compare performance of our method with unsupervised approaches of multi-shot SDALF~\cite{SDALF2010}, Color+LFDA~\cite{Pedagadi13}, 
Salience Match~\cite{Zhao13} (Salience), multi-shot extension of Fisher Vector descriptors~\cite{Ma12} (FV2D), 
3D Fisher descriptors around Flow Energy Profile (FEP) extrema~\cite{Liu15} (FV3D), Discriminatively Trained Viewpoint Invariant Dictionaries~\cite{Karanam15} (DVDL) 
and Fisher descriptors for spatio-temporal body-action units~\cite{Liu15} (STFV3D). Most of these approaches use HOG and/or Color as low-level feature descriptors, however, vary in how this information 
is combined to represent appearach of a person. 
As Table~\ref{tab:sota-prid-unsup} shows, our method significantly outperforms all these approaches. As we also use similar low-level feature descriptors, it is reasonable to argue that the 
improved performance is a consequence of improvement in signature representation. 

\begin{table}[bt]
     \begin{subtable}[b]{0.47\linewidth}
        \centering
	\begin{tabular}{|l|*{4}{c|}}
	\hline
	Method & r=1 & r=5 & r=10 & r=20 \\ \hline 
	Color+LFDA\cite{Pedagadi13} & 43.0 & 73.1 & 82.9 & 90.3 \\
	SDALF\cite{SDALF2010} & 5.2 & 20.7 & 32.0 & 47.9 \\  
	Salience\cite{Zhao13} & 25.8 & 43.6 & 52.6 & 62.0 \\  
	FV2D\cite{Ma12} & 33.6 & 64.0 & 76.3 & 86.0 \\  
	FV3D\cite{Liu15} & 38.7 & 71.0 & 80.6 & 90.3 \\  
	DVDL\cite{Karanam15} & 40.6 & 69.7 & 77.8 & 85.6 \\  
	STFV3D\cite{Liu15} & 42.1 & 71.9 & 84.4 & 91.6 \\  \hline
 	MCAM-LR & 50.9 & 79.4 & 87.2 & 94.7 \\ 
	MCAM-CRCS & 53.9 & 78.9 & 86.7 & 94.4 \\
 	MCAM-LR+CRCS & \textbf{58.9} & \textbf{83.9} & \textbf{93.3} & \textbf{96.9} \\ \hline 
	\end{tabular}
    \caption{Models without supervised learning}
    \label{tab:sota-prid-unsup}
    \end{subtable}
    ~
    \begin{subtable}[b]{0.5\linewidth}
        \centering
	\begin{tabular}{|l|*{4}{c|}}
	\hline
	Method & r=1 & r=5 & r=10 & r=20 \\ \hline 
	Color+DVR\cite{Wang2014} & 41.8 & 63.8 & 76.7 & 88.3 \\
	ColorLBP+DVR\cite{Wang2014} & 37.6 & 63.9 & 75.3 & 89.4 \\  
	ColorLBP+RSVM\cite{Wang2014} & 34.3 & 56.0 & 65.5 & 77.3 \\  
	DVR\cite{Wang2014} & 28.9 & 55.3 & 65.5 & 82.8 \\  
	DSVR\cite{Wang2016} & 40.0 & 71.7 & 84.5 & 92.2 \\  
	Salience+DVR\cite{Wang2014} & 41.7 & 64.5 & 77.5 & 88.8 \\  
	SDALF+DVR\cite{Wang2014} & 31.6 & 58.0 & 70.3 & 85.3 \\  
	STFV3D+KISSME\cite{Liu15} & \textbf{64.1} & \textbf{87.3} & 89.9 & 92.0 \\  \hline
 	MCAM-LR+CRCS & 58.9 & 83.9 & \textbf{93.3} & \textbf{96.9} \\ \hline 
	\end{tabular}
    \caption{Models with supervised learning}
    \label{tab:sota-prid-sup}
    \end{subtable}
		\vspace{-6mm}
  \caption{Comparison of MCAM with state-of-the-art on PRID2011 dataset using recognition rate (\%) at different ranks $r$.}
	\vspace{-8mm}
\end{table}

Next, we compare performance of our approach with supervised model learning approaches Discriminative Video Ranking~\cite{Wang2014} (DVR), Color features with DVR~\cite{Wang2014}, 
Color and LBP Features with DVR~\cite{Wang2014} (ColorLBP+DVR), Color and LBP Features with Rank SVM~\cite{Wang2014} (ColorLBP+RSVM), 
SDALF with DVR~\cite{Wang2014} (SDALF+DVR), Salience Matching with DVR~\cite{Wang2014} (Salience+DVR), 
Discriminative Selection in Video Ranking~\cite{Wang2016} (DSVR), and STFV3D+KISSME~\cite{Liu15}. 
The $rank-N$ recognition rates of these approaches and our method are given in Table~\ref{tab:sota-prid-sup}. 
With an exception to STFV3D+KISSME, our approach significantly outperforms all other methods. 
STFV3D+KISSME approach gives significantly superior performance for ranks less than 10, 
however, our method is better for higher ranks. It important to note that the performance gap between MCAM and STFV3D without metric learning is quite significant in favor of MCAM. 
The reason for MCAM to perform this well is that DVR based methods use FEP extrema as cue for multi-modality, which 
limits performance of metric learning. On the other hand, STFV3D uses body part information to regulate the FEP. This improves quality of signatures and hence it is able to learn color inconsistency 
between two cameras better. This strengthens our claim that quality of signatures dictates the upper bound on performance of supervised learning. 
Therefore, our method's performance is not an anomaly but a derivative of improved signature representation.

\noindent\textbf{iLIDS-VID:}
For fair comparison with other methods, we use experimental setup of~\cite{Wang2014}. Data is equally and randomly divided into train and test sets. 
Train set is then discarded. 
We first compare performance of our method against the same the unsupervised approaches used for comparison on 
PRID 2011 dataset,~\ie Color+LFDA~\cite{Pedagadi13}, multi-shot SDALF~\cite{SDALF2010}, 
Salience~\cite{Zhao13}, FV2D~\cite{Ma12}, FV3D~\cite{Liu15}, DVDL~\cite{Karanam15} and STFV3D~\cite{Liu15}. 
As Table~\ref{tab:sota-vid-unsup} shows, our method outperforms all approaches for ranks up to 15. Since, the underlying low-level features are similar among approaches, 
it is reasonable to attribute performance improvement to the use of MCAM for Re-ID task. 

\begin{table}[bt]
     \begin{subtable}[b]{0.47\linewidth}
        \centering
	\begin{tabular}{|l|*{4}{c|}}
	\hline
	Method & r=1 & r=5 & r=10 & r=20 \\ \hline 
	SDALF\cite{SDALF2010} & 5.1 & 19.0 & 27.1 & 37.9 \\  
	Salience\cite{Zhao13}~ & 10.2 & 24.8 & 35.5 & 52.9 \\  
	FV2D\cite{Ma12} & 18.2 & 35.6 & 49.2 & 63.8 \\  
	FV3D\cite{Liu15} & 25.3 & 54.0 & 68.3 & \textbf{87.7} \\  
	DVDL\cite{Karanam15} & 25.9 & 48.2 & 57.3 & 68.9 \\  
 	STFV3D\cite{Liu15} & 37.0 & 64.3 & \textbf{77.0} & 86.9 \\  \hline
 	MCAM-LR & 32.1 & 56.5 & 69.3 & 79.6 \\
	MCAM-CRCS & 34.3 & 60.4 & 71.5 & 81.7 \\
 	MCAM-LR+CRCS & \textbf{39.9} & \textbf{65.5} & \textbf{77.0} & 84.2 \\ \hline 
	\end{tabular}
    \caption{Models without supervised learning}
    \label{tab:sota-vid-unsup}
    \end{subtable}
    ~
    \begin{subtable}[b]{0.5\linewidth}
        \centering
	\begin{tabular}{|l|*{4}{c|}}
	\hline
	Method & r=1 & r=5 & r=10 & r=20 \\ \hline 
	MLF\cite{Zhao2014} & 11.7 & 29.1 & 40.3 & 53.4 \\  
	Color+RSVM\cite{Wang2014} & 16.4 & 37.3 & 48.5 & 62.6 \\ 
	ColorLBP+DVR\cite{Wang2014} & 32.7 & 56.5 & 67.0 & 77.4 \\  
	ColorLBP+RSVM\cite{Wang2014} & 20.0 & 44.0 & 52.7 & 68.0 \\  
	DVR\cite{Wang2014} & 23.3 & 42.4 & 55.3 & 68.6 \\  
	DSVR\cite{Wang2016} & 39.5 & 61.1 & 71.7 & 81.0 \\  
	MTL-LORAE\cite{Su15} & 43.0 & 60.1 & 70.3 & 85.3 \\ 
	STFV3D+KISSME\cite{Liu15} & \textbf{43.8} & \textbf{69.3} & \textbf{80.0} & \textbf{90.0} \\ \hline 
 	MCAM-LR+CRCS & 39.9 & 65.5 & 77.0 & 84.2 \\ \hline 
	\end{tabular}
    \caption{Models with supervised learning}
      \label{tab:sota-vid-sup}
    \end{subtable}
		\vspace{-6mm}
      \caption{Comparison of MCAM with state-of-the-art on iLIDS-VID dataset using recognition rate (\%) at different ranks $r$.}
		\vspace{-8mm}
\end{table}

In the end, we compare performance of our approach with same supervised learning approaches as used for comparison for PRID 2011~\ie DVR~\cite{Wang2014}, 
ColorLBP+DVR~\cite{Wang2014}, ColorLBP+RSVM~\cite{Wang2014}, 
Salience+DVR~\cite{Wang2014}, SDALF+DVR~\cite{Wang2014}, DSVR~\cite{Wang2016} and STFV3D+KISSME~\cite{Liu15}. In addition, we compare with 
 Mid-level Filters~\cite{Zhao2014} (MLF) and Multi-Task Learning with Low Rank Attribute Embeddding~\cite{Su15} (MTL-LORAE). 
Once again, our approach is only inferior to STFV3D+KISSME for ranks less than 20. 
However, unlike PRID 2011, effect of metric learning on STFV3D is relatively small because of better color consistency between cameras. Therefore, the amount of 
additional improvement from metric learning is related to the deviation of data from one camera to another. It should be noted that STFV3D and DVR require each track to be at least 21 frames long. 
Our method does not have this restriction. 

\vspace{-3mm}
\section{Conclusion}
\vspace{-2mm}
This paper addresses appearance based multi-shot person re-identification problem by proposing an extensible model to represent a person's appearance. 
To add robustness to pose, viewpoint and illumination changes, a person's appearance is explcitly modeled as a multi-modal feature density using GMMs. 
The strategy is to use variance of low level feature as a cue to discover different modes of appearance of a person, unlike earlier approaches which use either motion or viewpoint based cues. 
As the representation allows for a number of low level features to be integrated together, multiple appearance models of a person are learned, one for each feature, independently of others. 
This allows the representation to better preserve complementary nature of features. 
Furthermore, as the appearance models are represented as GMMs, this opens up new doors to measure similarity between two signatures using statistical distances, such as f-divergence. 
We chose to implement a derivative of divergence based metric and complemented it with dictionary coding based distance. There is a potential that proposed metric can be modified as a parametric 
measure with supervised learning, but we chose to avoid this for annotation cost. Even though different components of the approach and use of multiple featuers are not novel, their thoughtful assembly in a novel way yields significant 
performance improvement over competing multi-feature approaches.

We evaluated proposed representation on three challenging benchmark datasets (SAIVT-SoftBio, PRID2011, and iLIDS-VID) for multi-shot Re-ID using a set of complementary features 
to capture color, shape and texture information and outperformed state-of-the-art methods on all of them. 
Importantly, the matching approach does not rely on supervised metric learning; instead, the improvement in performance is a consequence of signature robustness to different artifacts in person's appearance 
across different cameras. This makes the proposed approach suitable for real-world Re-ID systems. 

\noindent\textbf{Acknowledgement:} The research leading to these results has received funding from the People Programme (Marie Curie Actions) of the 
European Union's Seventh Framework Programme FP7/2007-2013/ under REA grant agreement No. 324359.

\bibliographystyle{splncs}
\bibliography{biblo}
\end{document}